# Efficient Bi-manipulation using RGBD Multi-model Fusion based on Attention Mechanism

Jian Shen, Jiaxin Huang, Zhigong Song*
School of Mechanical Engineering, Jiangnan University, Wuxi 214122, China.
song_jnu@jiangnan.edu.cn

**Abstract** : Dual-arm robots have great application prospects in intelligent manufacturing due to their human-like structure when deployed with advanced intelligence algorithm. However, the previous visuomotor policy suffers from perception deficiencies in environments where features of images are impaired by the various conditions, such as abnormal lighting, occlusion and shadow etc. The Focal CVAE framework is proposed for RGB-D multi-modal data fusion to address this challenge. In this study, a mixed focal attention module is designed for the fusion of RGB images containing color features and depth images containing 3D shape and structure information. This module highlights the prominent local features and focuses on the relevance of RGB and depth via cross-attention. A saliency attention module is proposed to improve its computational efficiency, which is applied in the encoder and the decoder of the framework. We illustrate the effectiveness of the proposed method via extensive simulation and experiments. It's shown that the performances of bi-manipulation are all significantly improved in the four real-world tasks with lower computational cost. Besides, the robustness is validated through experiments under different scenarios where there is a perception deficiency problem, demonstrating the feasibility of the method.

*Index Terms* —Imitation Learning for robot and automation, Bimanual Manipulation, RGB-D Perception

## I. INTRODUCTION

With the rapid development of intelligent technologies [1], [2] and robotic equipment [3], robot manipulation has made significant progress in recent years due to an increasing demand in the real world. Traditional robotic equipment is usually employed in various practical scenarios, which feature standardized processes, repetitive tasks, and labor-intensive operations, such as welding [4] and component assembly [5] in the automotive industry, PCB soldering [6] and component assembly [7] in electronics , and additive manufacturing [8]. To perform these tasks, researchers have developed various classical control algorithms based on kinematics, including model predictive control [9], robust control [10], and sliding mode control [11]. Owing to the state-of-the-art control theory and upgraded hardware, robotechnology has been expanding to some nonstandard scenarios, such as surgical assistance [12], post-disaster rescue [13], deep-water exploration [14], and space station maintenance[15]. These tasks demand superior environmental adaptability, consequently, environmental sensing and perception are key components in control algorithms.

In these nonstandard scenes, classic control methods don't work due to the dependencies on predefined rules [16]. In order to enable the robot to perceive the environment, computer vision technology was applied in the field. By image processing and feature matching, the robot can localize and recognize potential objects. However, the method has limitations in the following scenarios: abrupt lighting changes, occlusion, and shadow, as well as overlapping objects. Under these circumstances, the effectiveness of object recognition and detection will be significantly impaired.

To deal with the above-mentioned issues, reinforcement learning is applied to endow the robot with environmental adaptability. Following this methodology, robots can interact with the environment to gain a better understanding and optimize their behavioral strategies through trial-and-error learning under the guidance of a proper reward function[16]. However, when it comes to high-dimensional perceptual data, the size of the state space will increase dramatically [17], which not only increases the computation cost, but also hinders the learning performance. This problem remained unresolved until DeepMind proposed DQN [17] in 2015, which is the first deep reinforcement learning algorithm, realizing stable and effective learning under high-dimensional perceptual inputs. OpenAI then proposed the policy-gradient-based algorithm PPO [18] in 2017, which significantly improved the sample efficiency and the stability during the training process. Haarnoja et al. proposed SAC [19], incorporating the principles of maximum entropy within the framework of Actor-Critic [20], which enhances the robustness of the learning process.

However, reinforcement learning tends to suffer from slow convergence due to its vast exploration space. In contrast, imitation learning provides a method to rapidly acquire skills by imitating expert behavior. Different from the trial-and-error process of reinforcement learning, imitation learning acquires decision-making strategies directly from expert performers, thus enabling robots to quickly adapt to diverse environments without an explicit reward function. For example, Behavior Cloning [21] learns the behavioral strategies of experts directly within the supervised learning paradigm. Inverse Reinforcement Learning [22] derives a reward function from the presented expert demonstration data, and the agent can perform under the guidance of this reward function. Generative Adversarial Imitation Learning [23] is inspired by the Generative Adversarial Networks (GAN) [24] in imitation learning, which learns effective strategies through the game-like confrontation between generators and discriminators.

Visual environment information is crucial for perception in imitation learning [25], [26], [27], [28]. However, it will suffer from visual perception deficiencies when faced with visually confusing environments [29], [30]. For instance, this includes challenges in poor lighting conditions, occlusions, and shadows, or backgrounds of similar color. Specifically, in underlit or overexposed conditions, an image's features can be significantly impaired, resulting in poor performance in feature extraction and object recognition. Shadows can be recognized as part of an object or mistaken for another object. Occlusion may cause the loss of key feature information. In addition, objects are difficult to distinguish from the background when their colors or textures are very similar. To improve the robot's performance in the above situations, some scholars have introduced depth images, which can provide 3D structure information for perception [33]. Through RGB-D fusion, the robot is endowed with comprehensive perceptual abilities that significantly improve performance in changing environments.

To address the problem of visual perception deficiency, a mixed focal attention module has been constructed within the framework of a CVAE-based imitation learning algorithm, which fuses RGB with depth information to provide richer environmental information for manipulation. Our main contributions are as follows:

(1) We propose an effective attention-based multi-modal fusion network that extracts the features of RGB and depth respectively, and then fuses environmental information using an efficient cross-attention module.

(2) We propose an efficient sparse attention module based on attention scores, which is applied to the encoder and decoder in the algorithm. It highlights the key frame of motion trajectory during the ranking of the attention coefficients, thereby reducing the computational cost.

(3) Through experiments in simulation and ablation studies, we demonstrate that the fusion of depth and RGB can effectively improve the performance of robots when encountering visual perception deficiencies. In addition, we collect 50 expert trajectories in four real-world tasks, respectively, through which we validate the robustness of our algorithm in the real world.

## II. RELATED WORKS

### A. Imitation Learning for Bimanual Manipulation

Dual-arm robots used to be expensive and their control policy is complex due to a high degree of freedom. Zhao et al. [31] presented a low-cost system that performs end-to-end imitation learning directly from real demonstrations. In addition, they developed an efficient algorithm, Action Chunking with Transformers [31], reducing the compounding-error in imitation learning. Shi [32] decomposed the demonstration into a minimal set of waypoints, which, when interpolated linearly, can approximate the trajectory, reducing the horizon of the learning problem and thus the errors compounded over time. Adopting the popular diffusion model as a generator, Chi et al.[33] offered a method that optimizes the gradient of the action policy using stochastic Langevin dynamics. By using probabilistic generative models, the robot policy can perfectly handle the multi-modal action distributions. These works can achieve superb performance in normal scenes but are unable to adapt to conditions with perception deficiencies due to relying solely on RGB information as input.

### B. Deep Multimodal Learning for robotics

Multimodal data enables a comprehensive perception of the environment in imitation learning. Du et al. [34] fused audio and vision in robot manipulation tasks, using LSTM (Long Short-Term Memory) to encode the history of visual and audio observations. This approach improved performance in the presence of visual occlusions, but the acoustic feedback tended to have some lag. Furthermore, Li [35] enabled robots to solve complex manipulation tasks by fusing visual, auditory, and tactile perception and analyzing attention scores of all three sensory modalities in different task phases. The necessity and effectiveness of multimodal fusion in robot manipulation tasks is verified in these works. The emergence of large language models in recent years has driven the rapid development of multimodal embodied intelligence in robotics [36], [37], [38]. These methods not only endow robots with autonomous capabilities, but also significantly improve their generalization and adaptability in diverse environments. This advancement is a breakthrough in the field of robotics, but it suffers from an extremely high computational cost. When the model is deployed on robot systems, such computations can result in severe latency.

### C. Computational optimization based on sparse attention mechanism

As multimodal data increases the computational complexity of robot manipulation tasks, traditional attention mechanisms are difficult to apply to an edge device on some specific occasions. To solve this problem, researchers have explored many sparse attention mechanisms, aiming to reduce the computational cost while maintaining the performance. Sliding window attention [39] divides the input into multiple smaller windows or chunks, and performs attention within these localized windows. Axial Attention [40] proposes a position-sensitive self-attention design, optimizing the computation by factorizing 2D self-attention into two 1D self-attentions. Compared to the above methods, Deformable Attention [41], [42] only attends to a small set of key sampling points around a reference, thus enabling the module to focus on relevant regions and capture more informative features. Informer [43] reduces the sequence length by Temporal Down-Sampling and performs attention with probsparse attention, optimizing the efficiency. As these approaches have achieved great performance in computer vision, they offer great potential in robot control.

To address the above issues, we propose the Focal-CVAE. Our method utilizes mixed focal attention to improve the representation of features in the environment. In addition, salient attention is proposed to simplify feature extraction by computing the saliency coefficient of the action frame, ensuring the real-time performance of the algorithm.

## III. METHODOLOGY

### A. Structure Overview of Focal-CVAE

In this work, the Focal-CVAE framework is proposed to solve the problem of perception deficiencies encountered in real-world scenarios for manipulation tasks. The framework consists of an encoder and a decoder based on saliency attention, a Mixed Focal Attention module (MFA) for feature extraction of environmental visual information, and a linear projection layer for extracting features of proprioception. Its network architecture is shown in Fig. 1.

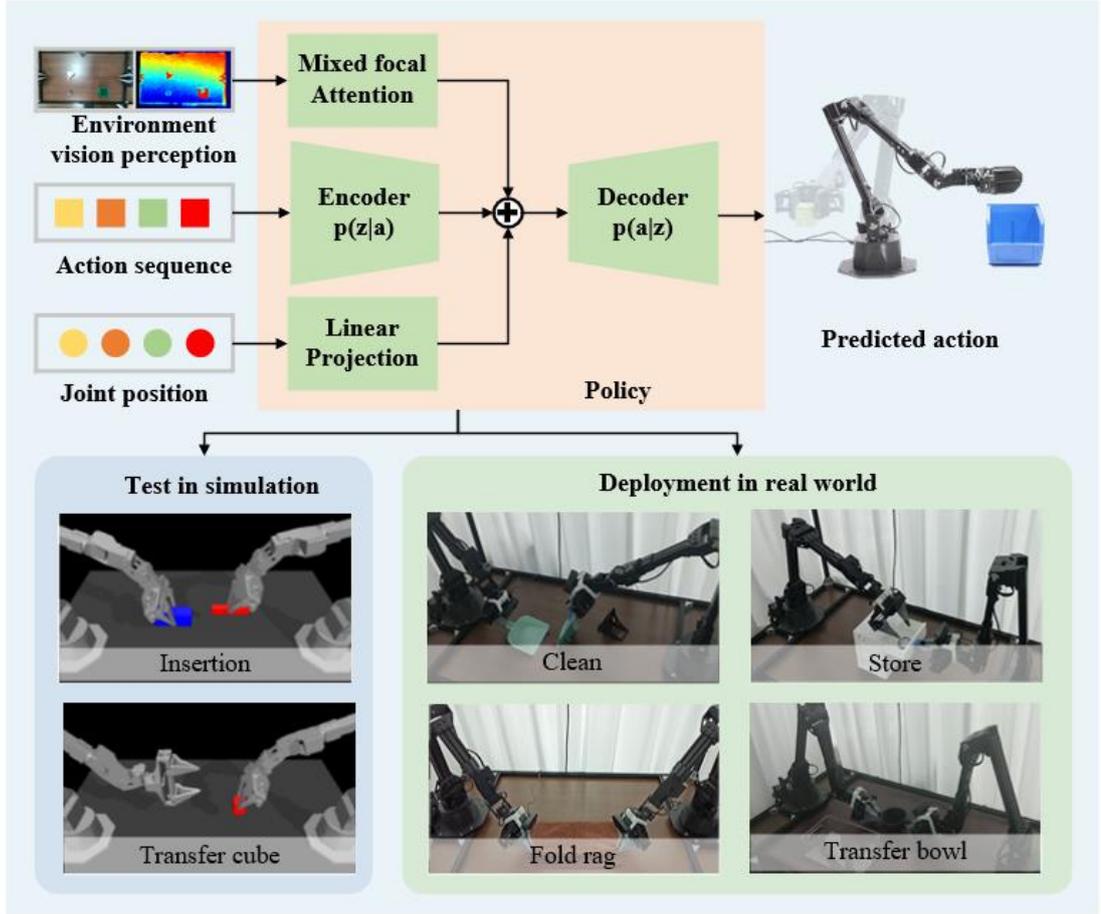

Fig. 1. The upper part is the architecture of the proposed Focal-CVAE, which has an encoder, a decoder, a mixed focal attention module and a linear projection module for extracting visual and proprioceptive feature respectively. Besides we show the learned skills from two tasks in simulation and four tasks in real world in the lower part.

In the MFA module, an effective sparse attention mechanism is introduced for image feature processing, where pixel points with the highest relevance are adaptively selected by the Focal Net. Subsequently, RGB and depth images are fully integrated using cross-attention. Concurrently, the proprioceptive data from the dual-arm system is processed in the linear projection module. Specifically, joint positions and velocities are taken as inputs, and kinematic features are extracted through multiple dense layers. The Focal-CVAE utilizes visual and kinematic information obtained from the MFA and linear projection as additional conditional inputs to the CVAE (Conditional

Variational Auto Encoder). This enhancement enables the robot to better understand dynamic behaviors in the environment, thereby contributing to improved collaborative manipulation between each arm.

Furthermore, we propose saliency attention aimed at improving the feature extraction of sequence data. Unlike hand-crafted sparse attention patterns, the feature is adaptively sparsified by ranking the attention scores. In this way, we optimize the structure of the encoder and the decoder, thereby improving computational efficiency.

*B. Mixed focus attention*

In order to improve the performance of manipulation tasks when the robot encounters perception deficiencies, we design a mixed focal attention module, which aims to fuse the color information from RGB images and the stereoscopic information from depth images, thereby obtaining comprehensive environmental information. With this information, the robustness of the policy can be enhanced in various scenarios. The specific structure is illustrated in Fig. 2.

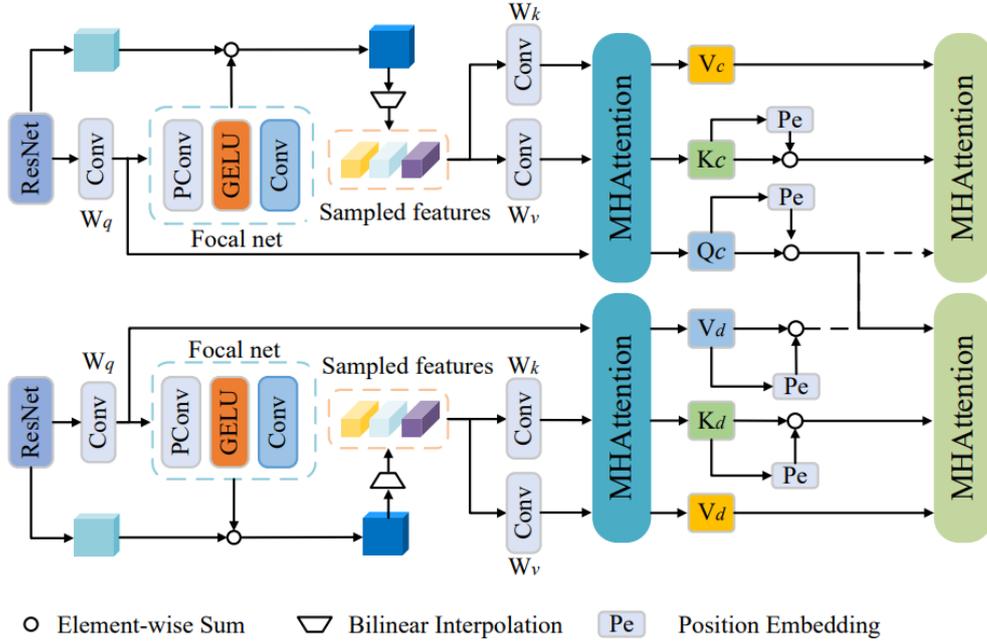

Fig. 2. Detail architecture of the mixed focal attention module.

In the MFA module, we adopt Resnet as the backbone, which extracts features of RGB and depth respectively and simultaneously compresses the size of the feature map. In the following attention module, every pixel doesn't need to calculate an attention score with other pixels, but only with several relevant points located by offsets. The offsets are learned from the queries by the Focal Network. The model can not only reduce memory and computation costs but also avoid irrelevant parts that are beyond the region of interest.

More specifically, a uniform grid of points will be generated as a reference. The grid size is calculated from the input feature map size by the factor $r$. Given the input feature map $\in R^{H\times W\times C}$, the grid of point $\in R^{HG\times WG\times 2}$, $HG = H/r$, $WG=W/r$. To get the offset of every reference point, the focal net takes the $W_q$ as input. The focal net is consists of three layers: PConv[44], GELU and a 1×1Conv layer. Compared to conventional deformable attention, it replaces DWC (Depthwise Separable Convolution) with PConv. This module extracts the spatial features in channel dimension, which significantly improves the efficiency of the feature maps. Then, after a GELU activation layer and a 1×1 Conv layer, the 2D offsets can be calculated. We sample the corresponding feature according to the offsets, and we can obtain a focused feature that is smaller than the original feature

map. Finally, we obtain the key and value matrices following the projection matrices:

$$\begin{aligned} q = xW_q, \tilde{k} = \tilde{x}W_k, \tilde{v} = \tilde{x}W_v \\ \Delta p = \theta_{offset}(q), \tilde{x} = f(x; p + \Delta p) \end{aligned} \quad (1)$$

$k$, $v$ represents the embedding of key and value. And then perform multi-head attention with relative position embedding as following formulation:

$$z^m = \sigma\left(q^m \tilde{k}^{m\top} / \sqrt{d} + \phi(\hat{B}; R)\right)\tilde{v}^m \quad (2)$$

where $\phi(\hat{B}; R) \in \mathrm{R}^{HW \times HGWG}$ correspond to the position embedding, and $Zm$ represents the output of the $m$-th attention head. After a self-attention layer, position embedding is added to the Q and K matrices, and then cross-attention is performed on both features. Meanwhile, to aid in understanding the relevance of positions in sequence features, we add position embedding to the second multi-head attention module.

In the MFA module, computational complexity is reduced through focus attention. Meanwhile, we filter out redundant features that may be present in dense attention models. Compared to the attention mechanisms used in robot manipulation, our method effectively models the relationships among tokens under the guidance of important regions in the feature maps.

*C.  VAE based on saliency attention*

In the encoder of the CVAE, the efficiency of feature extraction for action sequences is crucial. In previous imitation learning algorithms, only the current action state is input, failing to describe the dynamic process through a discrete action frame alone. In order to capture the continuous motion pattern from the trajectory, we attempt to take a sequence of actions within a time slice as the input, but this inevitably increases the computational cost. To balance this extra complexity, we propose using saliency attention for computational optimization of action sequences. Specifically, we select the top-u queries under the sparsity measurement and regard them as the key frames of this trajectory. Then, the attention distribution is calculated by multiplying the *Q* and *K* matrices. When this distribution of a certain feature is more similar to the uniform distribution, it indicates that the difference in relevance of the feature with others is small; in other words, the feature is non-salient. On the other hand, a larger variation in the distribution indicates that the feature is significant. Through saliency measurement, we can select the key frames of a sequence, allowing for the extraction of high-dimensional action features. Furthermore, this improves efficiency and ensures real-time performance. The process of saliency measurement is as follows:

For each $q_i \in \mathrm{R}^d$ and every $k_j \in \mathrm{R}^d$ in set $K$, we calculate the distribution of their product. To simplify the process, we take the difference between the maximum and average values as the saliency measurement, as follows:

$$\bar{M}(q_i, K) = \max_j \left\{ \frac{q_i k_j^\top}{\sqrt{d}} \right\} - \frac{1}{L_K} \sum_{j=1}^{L_K} \frac{q_i k_j^\top}{\sqrt{d}} \quad (3)$$

$q_i$ is the $i$-th vector in query, $k_j$ is the $j$-th vector in key. $D$ is square root of the dimension of the feature, $L_k$ is the size of key matrix. The top-*u* queries are selected by the $M$ value as $Q'$ and then we perform attention on this query. As for the remaining part $\bar{Q}$, its features can be replaced by average value of *V* matrix. After cascading *N* such attention modules, the sequence features are extracted. Finally, after embedding layer the latent variable $Z$ is obtained.

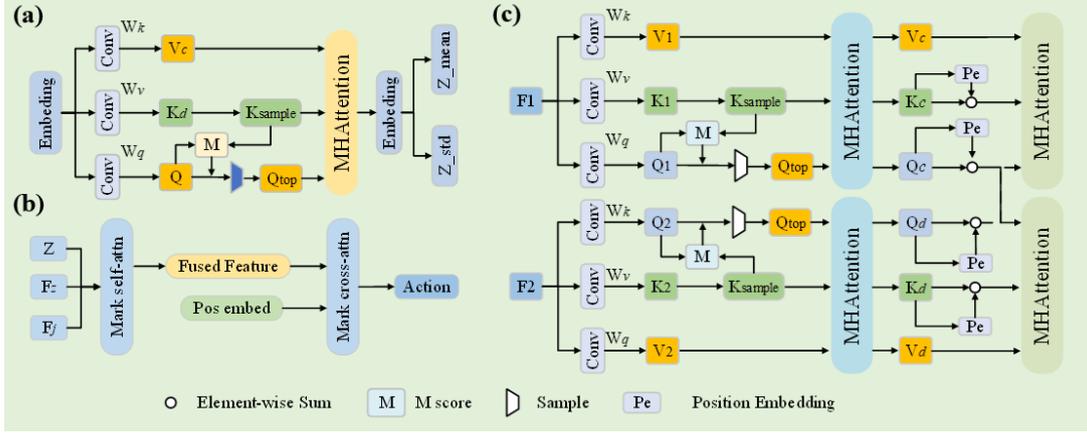

Fig. 3. (a) Detail architecture of encoder based on saliency attention. (b) Frame of decoder based on saliency attention. (c) Detail architecture of encoder based on saliency attention.

The decoder takes latent variable *z* from encoder, the visual feature *Fv* from the MFA and the proprioceptive feature *Fp* from the linear mapping layer as the inputs, and then concatenates these features using salient attention. Finally, the predicted action is generated via the Saliency Decoder.

In this way, the SCS (Saliency Cross-Attention) module enables the model to capture the salient features in the action sequences, thereby gaining a better understanding of the global action state. In data-dense occasions, it is effective in ensuring real-time performance.

### D. The loss function

In this framework, the loss function consists of reconstruction loss and regression loss.

In the training process, the reconstruction loss measures the similarity between the output and the input. By minimizing the reconstruction loss, the predicted actions can be made as close as possible to the original input. The formula is as follows:

$$\mathcal{L}_{reconst} = MSE(\hat{a}_{t:t+k}, a_{t:t+k}) \tag{4}$$

Where $a_t$ is the action at timestep *t*, $\hat{a}_t$ is the prediction action at timestep *t*.

The regression loss ensures that the latent space is similar to a standard Gaussian distribution, which has great characteristics. By minimizing the Kullback-Leibler (KL) divergence between the distribution of latent variables and the prior distribution, the latent space can converge to a normal Gaussian distribution. This process, known as "regularization," prevents model overfitting and helps to improve the model's generalization.

$$\mathcal{L}_{reg} = D_{KL}\left(q_\phi(z \mid a_{t:t+k}, \overline{o}_t) \parallel \mathcal{N}(0, I)\right) \tag{5}$$

Where $\overline{o}_t$ is the observation at timestep *t*, $a_t$ is the action at timestep *t*, *z* is the latent variable, $q_\phi$ is the decoder of the CVAE, $N(0, I)$ is a standard Gaussian distribution. The overall loss function is a combination of regression loss and reconstruction loss to generate representative latent variable. Therefore, the loss function is as follows:

$$\mathcal{L} = \lambda_1 D_{KL}\left(q_\phi(z \mid a_{t:t+k}, \overline{o}_t) \parallel \mathcal{N}(0, I)\right) + \lambda_2 MSE(\hat{a}_{t:t+k}, a_{t:t+k}) \tag{6}$$

Where $\lambda_1$、$\lambda_2$ is the coefficient of the regression loss and reconstruction loss in the Focal-CVAE.

# IV. EXPERIMENTAL RESULTS

## A. Experiment setup

We adopt two bimanual manipulation tasks in simulation: transferring cubes and inserting. In the real environment, we designed four bimanual manipulation tasks: transferring bowls, cleaning the table, folding rags, and storing items. These tasks were also performed under normal lighting and shadow conditions:

Table 1 Summary of Task-Specific Datasets and Phases for Robotic Manipulation.

| Dataset | Phase1 | Phase2 | Phase3 | length | num | Total size |
|---|---|---|---|---|---|---|
| Transfer cube (sim) | grasp | lift | transfer | 600 | 50 | 36.87G |
| Insertion(sim) | grasp | contact | insert | 600 | 50 | 36.87G |
| Transfer bowl (real) | touched | lift | place | 600 | 50 | 135G |
| Clean the table (real) | grasp | approach | sweep | 600 | 50 | 135G |
| Fold rag (real) | touched | lift | fold | 600 | 50 | 135G |
| Store things (real) | open | grasp | place | 600 | 50 | 135G |

The software and hardware for training are as follows: The experiment is conducted on Ubuntu 20.04, and we use Pytorch as the framework. The device is configured with an Intel(R) Xeon(R) Platinum 8175M CPU @ 2.50GHz and an NVIDIA GeForce RTX 3060 equipped with 12GB of RAM.

## B. Performance evaluation

There are two simple tasks in the simulation: transferring a cube and insertion. This experiment is conducted under normal and dim conditions, respectively.

Table 2 Success rate (%) for the 2 baseline tasks in normal scene.

| Normal scene | Cube Transfer | | | Bimanual Insertion | | |
|---|---|---|---|---|---|---|
| | Touched | Lifted | Transfer | Grasp | Contact | Insert |
| ACT | 97 | 90 | 86 | 93 | 90 | 32 |
| ours | 96 | 94 | **94** | 100 | 98 | **58** |

Table 3 Success rate (%) for the 2 baseline tasks in background color.

| Background color | Cube Transfer | | | Bimanual Insertion | | |
|---|---|---|---|---|---|---|
| | Touched | Lifted | Touched | Lifted | Contact | Insert |
| ACT | 86 | 86 | 82 | 76 | 48 | 12 |
| Ours | 98 | 92 | **90** | 94 | 44 | **40** |

In this experiment, the light source was removed in the simulation, making the potential object and table colors nearly the same. This enhances the difficulty of recognition. The outcomes show that the success rate of ACT in the insertion task is greatly reduced, while our model still maintains a high success rate.

For the real experiment, we set up a tele-operation system that includes two WidowX 250 arms and two ViperX 300 arms. When we move the leader arm, the joint angles are mapped to the follower arm. Images are recorded by the cameras mounted on the end effector and an environmental camera above the table. All action and image data are saved in ROS (Robot Operating System). Based on these data, the robot can complete tasks as shown in Fig. 4.

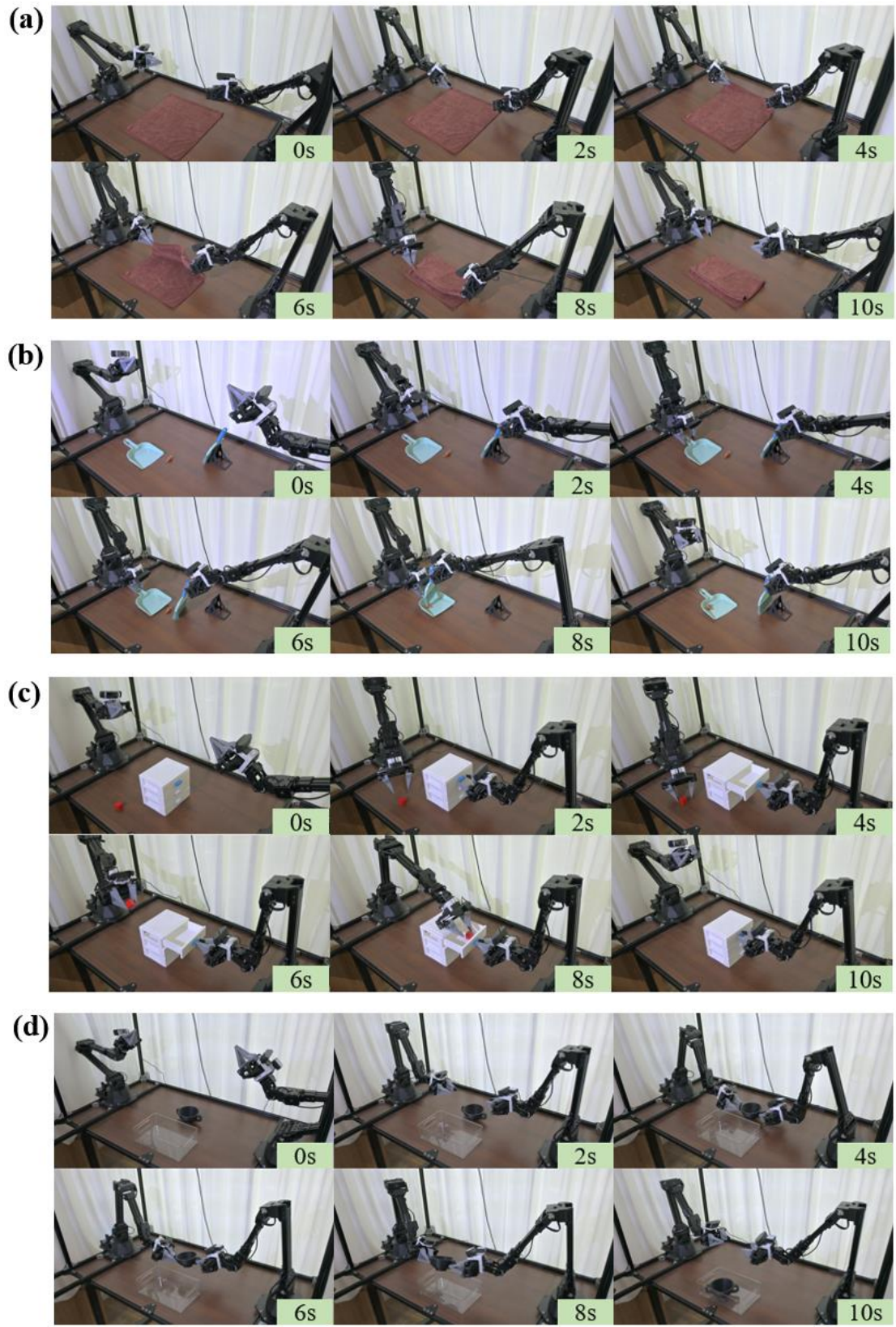

Fig. 4 Sample frames from the rollouts from our model on the follow real robot experiments.

(a): Fold rag. (b): Clean the. table (c): Store things. (d): Transfer bowl.

The outcome is shown in Table 4 as follows.

Table 4 Success rate (%) for the remaining 4 real-world tasks.

| Tasks | Methods | Normal | | | Under the shadow | | |
|---|---|---|---|---|---|---|---|
| | | Phase1 | Phase2 | Phase3 | Phase1 | Phase2 | Phase3 |
| Fold rag | ACT | 100 | 96.7 | 96.7 | 100 | 93.3 | 86.7 |
| | Ours | 100 | 97.1 | 97.1 | 100 | 96.7 | 93.3 |
| Transfer bowl | ACT | 96.9 | 92.5 | 92.5 | 90.6 | 81.3 | 81.3 |
| | Ours | 100 | 93.7 | 93.7 | 96.9 | 90.3 | 90.3 |
| Clean the table | ACT | 90.0 | 83.3 | 76.7 | 83.8 | 80.6 | 74.2 |
| | Ours | 90.0 | 86.7 | 83.3 | 87.1 | 83.9 | 77.4 |
| Store things | ACT | 100 | 87.5 | 68.4 | 100 | 81.3 | 62.5 |
| | Ours | 100 | 93.8 | 75.8 | 100 | 93.8 | 74.1 |

In task A, the position of the rag needs to be detected, and the two corners of the rag recognized. After the two arms approach and clamp its two corners, they need to coordinate to grasp the rag and fold it. The difficulty lies in the manipulation of deformable materials, which requires a high degree of coordination between the arms. Under dim light, the success rate of ACT is 86.7%, a decrease of 10% compared to 96.7% under normal light, indicating that the task is more sensitive to light. Meanwhile, the success rate of our method decreases from 97.1% to 93.3%, a smaller decrease of 3.8%, demonstrating stronger robustness.

In task B, the locations of the broom, dustpan, and clutter need to be recognized. The two arms must grasp the broom and the dustpan separately and use the broom to sweep the clutter on the desktop into the dustpan. The difficulty involves using tools, controlling the relative positions of the broom, dustpan, and clutter, and performing fine manipulations. In dim light, our algorithm improves the success rate by 9% compared to ACT, demonstrating better grasping accuracy during the tool pickup phase.

In task C, the locations of the storage box and the desktop sundries need to be recognized. The two arms must open the drawer of the storage box, grab the sundries from the desktop, and place them into the drawer, before finally closing it. The main difficulty lies in recognizing small objects. In the grasping phase under dim light, our algorithm shows a 3.3% improvement in performance compared to ACT.

In task D, the positions of the plastic bowl and the transparent container need to be recognized. The two arms must simultaneously grasp both sides of the plastic bowl, lift it, and smoothly place it into the transparent container beside it. The difficulty involves recognizing the transparent container and coordinating the grasping of both sides of the bowl. Under dim light, the grasping success rate of ACT decreased by 6.2% compared to that under normal lighting. Meanwhile, the grasping accuracy of our method was unaffected and continued to maintain a high success rate.

*C. Ablation Study*

The ablation experiments are performed on two basic tasks in a simulation, comparing the success rates of the task sub-phases in different environments. The best results are marked in bold in Table 5. In the tasks under normal conditions, the baseline achieved success rates of 78% and 31%. When the MFA and SAT (Saliency Attention Transformer) were added, the success rates improved. This shows that the proposed Focal-VAE realizes a better fusion of data. Specifically, the method achieves a 16% gain in task success for transferring cubes and a 27% increase for part insertion. However, when potential objects and the background share the same color, relying solely on RGB cannot provide accurate position information, resulting in the failure of the baseline. With the fusion of depth information, a higher success rate can be achieved, although it is much lower than under normal conditions. However, with the MFA and SAT modules, significant improvements can be achieved in the pass block task (18%) and the part assembly task (28%).

Table 5 Success rate (%) for the ablation under normal scene.

| Normal scene | Cube Transfer | | | Bimanual Insertion | | |
|---|---|---|---|---|---|---|
| | Touched | Lifted | Transfer | Grasp | Contact | Insert |
| baseline | 99 | 84 | 78 | 99 | 88 | 31 |
| baseline + MFA | 95 | 86 | 81 | 96 | 86 | 52 |
| baseline + SAT | 96 | 87 | 80 | 97 | 84 | 48 |
| baseline + depth | 96 | 96 | 92 | 96 | 82 | 34 |
| Focal-CVAE (ours) | 96 | 94 | **94** | 100 | 98 | **58** |

Table 6 Success rate (%) for the ablation under background color scene.

| Background color | Cube Transfer | | | Bimanual Insertion | | |
|---|---|---|---|---|---|---|
| | Touched | Lifted | Transfer | Grasp | Contact | Insert |
| baseline | 0 | 0 | 0 | 0 | 0 | 0 |
| baseline + MFA | 0 | 0 | 0 | 0 | 0 | 0 |
| baseline + SAT | 0 | 0 | 0 | 0 | 0 | 0 |
| baseline + depth | 86 | 86 | 72 | 76 | 48 | 12 |
| Focal-CVAE (ours) | 98 | 92 | **90** | 94 | 44 | **40** |

Table 7 Comparison results on real experiment. The latency results are tested on NVIDIA GeForce RTX 3060

| Methods | Params(M) | FLOPs(G) | Latency(ms) | Average successful rate(%) |
|---|---|---|---|---|
| baseline + depth | 90 | 1261 | 8.17 | 48.25 |
| Focal-CVAE (ours) | 113 | 589 | 5.21 | 70.5 |

### D. Attention score distribution

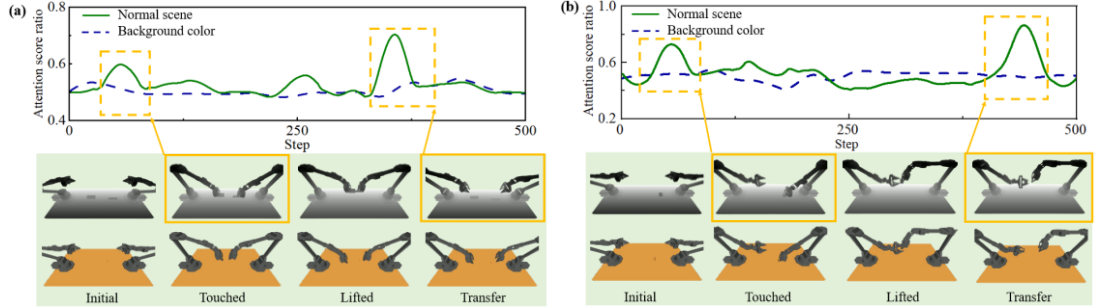

Fig. 5. Visualization of the aggregated attention scores for each modality as the robot completes following tasks (a): insertion. (b): transfer cube.

After the fusion of both RGB and depth information in Focal-VAE, the success rate of the task is significantly improved. To further analyze the proportion of the weight of a certain modality at a certain stage, attention score analysis is conducted. The two curves in the figure represent the percentages of depth and RGB weights. In the regular situation, the weights of both depth and color information are around 50% each. When faced with perception deficiency, there are two peaks in the depth weight, occurring in the grasping and passing phases. At these times, the object and the background have similar colors. Only the depth map can locate the potential object. In the second stage, the agent focuses more on the position of the end effector when depth information is less significant. It can be concluded that perception deficiencies can be effectively addressed by fusing depth information.

## V. CONCLUSION

In this work, we propose Focal-CVAE, an imitation learning framework for bi-manipulation. It aims to address visual perception deficiencies in complex scenarios and to enhance the efficiency of manipulation tasks. Through the use of mixed focus attention, RGB and depth features can be more effectively fused, providing richer environmental information for manipulation and further enhancing the adaptability of the algorithm. Additionally, saliency attention has been integrated into the encoder and decoder to simplify the computation of long sequence data, thereby further enhancing computational efficiency and ensuring the real-time performance of the algorithm. In future research, to optimize human-computer collaboration further, emphasis will be placed on user-friendly interaction interfaces, such as Augmented Reality (AR) and Virtual Reality (VR), which provide visual feedback to the operator and facilitate ease of use for non-expert users.